\documentclass[sigconf, nonacm]{acmart}

\def\BibTeX{{\rm B\kern-.05em{\sc i\kern-.025em b}\kern-.08emT\kern-.1667em\lower.7ex\hbox{E}\kern-.125emX}}
    
\copyrightyear{2025} 
\acmYear{2025} 
\setcopyright{rightsretained} 
\acmConference[CIKM '25]{The 31st ACM International Conference on Advances in Geographic Information Systems}{November 13--16, 2023}{Hamburg, Germany}
\acmBooktitle{The 31st ACM International Conference on Advances in Geographic Information Systems (CIKM '25), November 13--16, 2023, Hamburg, Germany}\acmDOI{10.1145/3589132.3625587}
\acmISBN{979-8-4007-0168-9/23/11}

\usepackage[utf8]{inputenc}
\usepackage[algo2e,linesnumbered,ruled,vlined]{algorithm2e}
\usepackage{graphicx}
\usepackage{subcaption}
\usepackage{float}
\usepackage[show]{chato-notes}
\usepackage{booktabs}
\usepackage{cancel}
\usepackage{amsmath}
\usepackage{enumitem}
\usepackage{comment}
\usepackage{multirow}
\usepackage{stfloats}

\begin{document}

\title[Human Mobility Datasets Enriched With Contextual and Social Dimensions]
{Human Mobility Datasets Enriched\\With Contextual and Social Dimensions}

\author{Chiara Pugliese}
\affiliation{%
  \institution{IIT-CNR \country{Pisa, Italy}}
}
\email{chiara.pugliese@iit.cnr.it}

\author{Francesco Lettich}
\affiliation{%
  \institution{ISTI-CNR \country{Pisa, Italy}}
}
\email{francesco.lettich@isti.cnr.it}

\author{Guido Rocchietti}
\affiliation{%
  \institution{ISTI-CNR \country{Pisa, Italy}}
}
\email{guido.rocchietti@isti.cnr.it}

\author{Chiara Renso}
\affiliation{%
  \institution{ISTI-CNR \country{Pisa, Italy}}
}
\email{chiara.renso@isti.cnr.it}

\author{Fabio Pinelli}
\affiliation{%
  \institution{IMT Lucca \country{Lucca, Italy}}
}
\email{fabio.pinelli@imtlucca.it}

\begin{abstract}
In this resource paper, we present two publicly available datasets of semantically enriched human trajectories, together with the pipeline to build them. The trajectories are publicly available GPS traces retrieved from OpenStreetMap. Each dataset includes contextual layers such as stops, moves, points of interest (POIs), inferred transportation modes, and weather data. A novel semantic feature is the inclusion of synthetic, realistic social media posts generated by Large Language Models (LLMs), enabling multimodal and semantic mobility analysis. The datasets are available in both tabular and Resource Description Framework (RDF) formats, supporting semantic reasoning and FAIR data practices. They cover two structurally distinct, large cities: Paris and New York. Our open source reproducible pipeline allows for dataset customization, while the datasets support research tasks such as behavior modeling, mobility prediction, knowledge graph construction, and LLM-based applications. To our knowledge, our resource is the first to combine real-world movement, structured semantic enrichment, LLM-generated text, and semantic web compatibility in a reusable framework.
\end{abstract}



\maketitle
\section{Introduction}
\label{sec: intro}

The widespread adoption of location-aware devices and location-based social media has led to the generation of vast amounts of tracking data. These data typically capture the time-stamped positions of moving entities, often enriched with contextual semantic information, e.g., visited points of interest (POIs), means of transportation, and weather conditions. However, despite the growing volume of such data produced by location service providers, the research community continues to face challenges in accessing publicly available semantically enriched human mobility datasets. This scarcity arises from a combination of proprietary ownership, commercial restrictions, and strict privacy regulations (e.g., the European GDPR). Additionally, each added semantic layer increases the risk of re-identification \cite{Gomes24,Monreale23}
and the complexity of data integration, making public release both difficult and sensitive.
The trajectory datasets available in the community often suffer from limitations such as being outdated or lacking semantic layers (e.g., Microsoft’s Geolife \cite{zheng2011geolife}, Porto Taxi dataset \cite{taxiporto2015}), or rely on spatiotemporal sparse check-ins with limited contextual information (e.g., Foursquare NYC/Tokyo \cite{yang2019revisiting}). To address these gaps, some researchers have turned to fully synthetic simulators \cite{AndreasPatterns24, GoceData24, KappSurvey24, matsim2016}, which, however, come with significant limitations in realism and applicability.

Despite these obstacles, we argue that semantically enriched trajectory datasets are critically important: they can provide deeper insights into mobility patterns and behaviors, serve as benchmarks for evaluating algorithms, and support the development of advanced mobility analytics, including the use of Large Language Models (LLMs) and spatial foundation models.
This resource paper introduces two publicly available datasets, along with the methodology and tools used to generate them. Our original contributions include:

\begin{itemize}[leftmargin=1em]
\item Two human mobility datasets, publicly released on Zenodo \cite{zenodo_dataset2025}, derived from real GPS trajectories voluntarily shared via OpenStreetMap (OSM) \cite{osm}, enriched with different semantic layers, including raw contextual information (weather conditions) and inferred attributes (stops, moves, POIs, and transportation means).

\item Synthetic social media data, generated using a carefully instructed LLM, that realistically simulates user posts associated with the movements of trajectories.

\item RDF (Resource Description Framework) representations of the datasets, in addition to the tabular ones, enabling 
semantic reasoning and LLMs integration. Furthermore, this format aligns with the FAIR (Findable, Accessible, Interoperable, Reusable) principles \cite{fair_principles}.

\item A reproducible and transparent pipeline for dataset creation, accompanied by publicly available code \cite{matdataset_github_2025}, allowing practitioners to generate customized versions of the datasets.

\item The two datasets can also be used incrementally: from the raw trajectories with no semantic aspects, to inferred semantic dimensions, to the fully enriched datasets that include realistic LLM-generated social media posts.
\end{itemize}
This resource paper aims to support researchers in mobility and knowledge management domains, enabling the experimentation, validation, and testing of novel methods for behavior modeling, semantic enrichment, multimodal mobility analysis, and urban knowledge discovery; as detailed in Section \ref{sec: task desc}, the versatility of our resource indeed addresses a wide range of research areas.

\section{Methodology and 
Datasets}
\label{sec: datasets}

\begin{figure*}[]
    \includegraphics[width=0.7\linewidth]{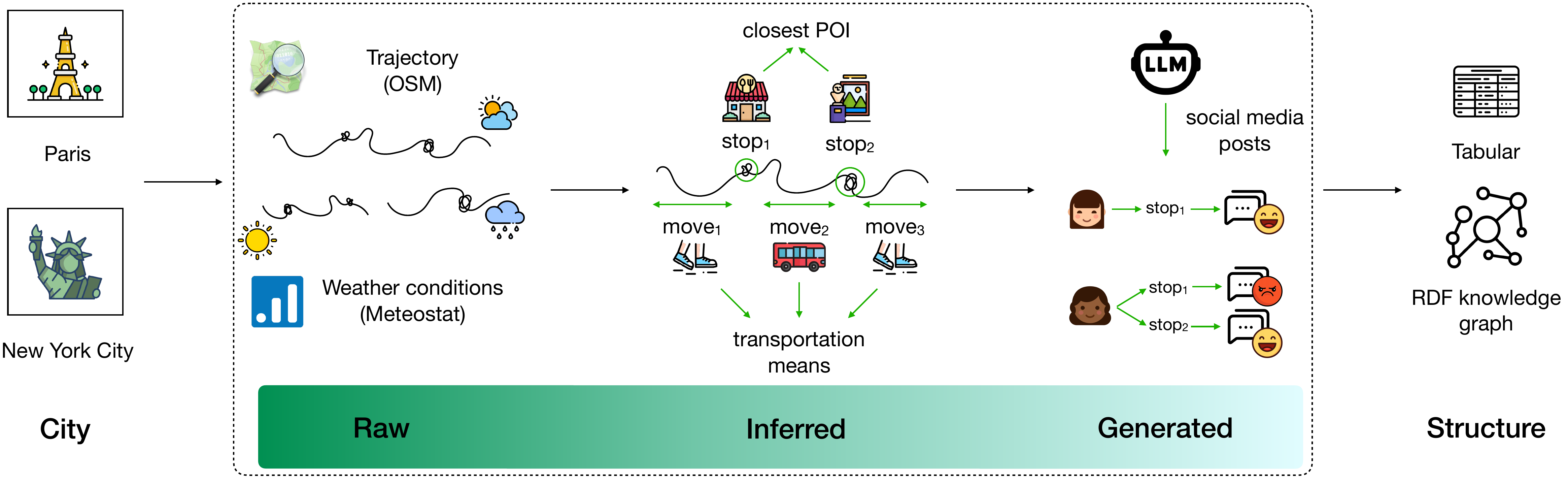}
    \caption{The semantic trajectories construction pipeline}
    \label{fig:schema}
\end{figure*}

The methodology to create the semantically enriched trajectories datasets is schematized in Figure~\ref{fig:schema}. The base layer of the figure represents the underlying data sources, specifically: (1) real-world raw data collected from OpenStreetMap \cite{osm} and Meteostat \cite{meteostat}; (2) the inferred mobility patterns (including stops, moves, visited POIs, and transportation means) extracted through a combination of algorithms and heuristics; (3) synthetic social media posts generated through a large language model (LLM) to simulate realistic content generated by users during their movements. The final output consists of datasets represented in both tabular and RDF formats.

\subsection{Data collection}
\label{sec: data collection}

The first step involves collecting the data to build the two datasets: raw GPS trajectories, POIs, and weather conditions.

\begin{figure}[h]
    \centering
    \includegraphics[width=0.485\columnwidth]{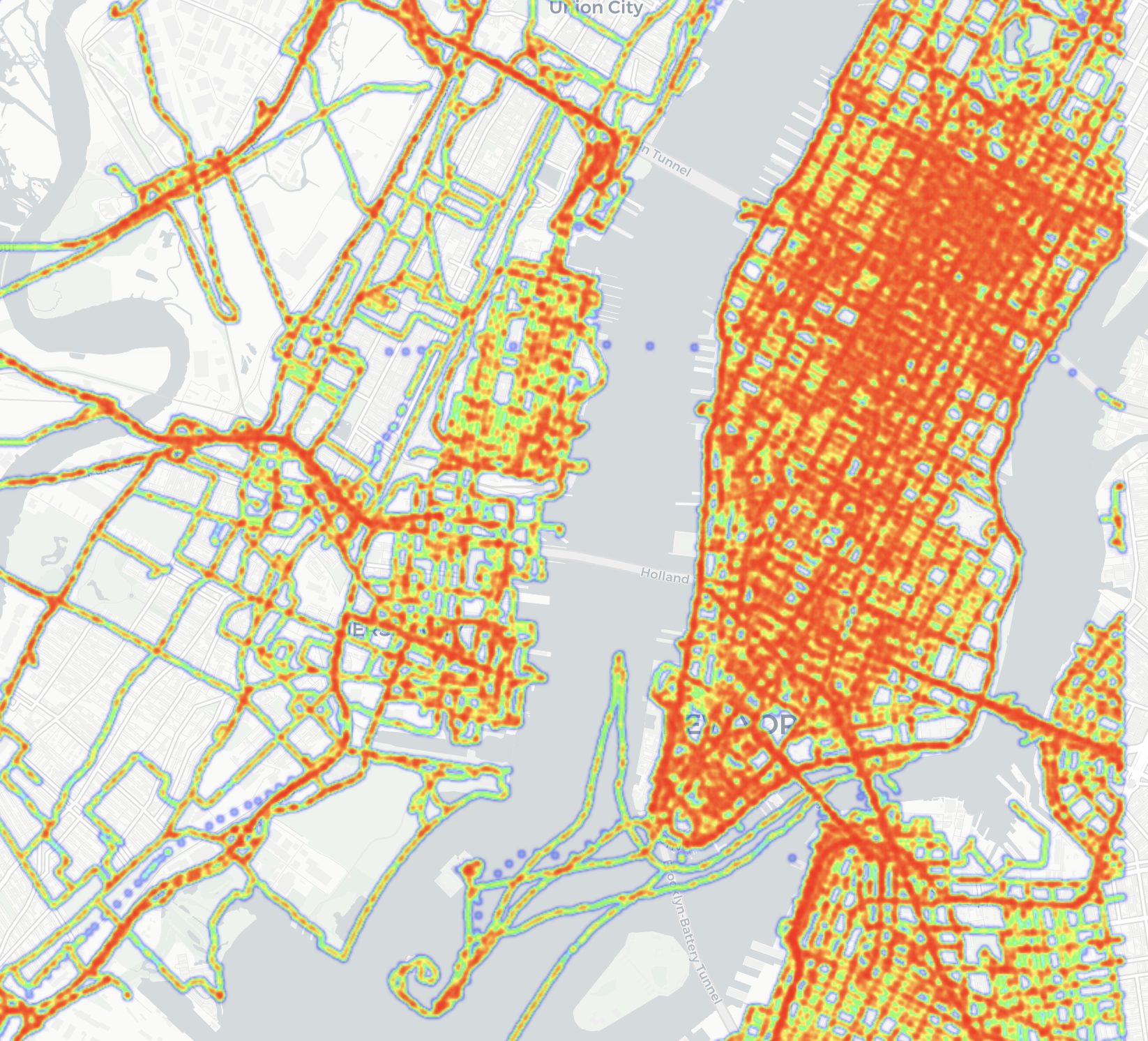}
    \hspace{0.1em}
    \includegraphics[width=0.485\columnwidth]{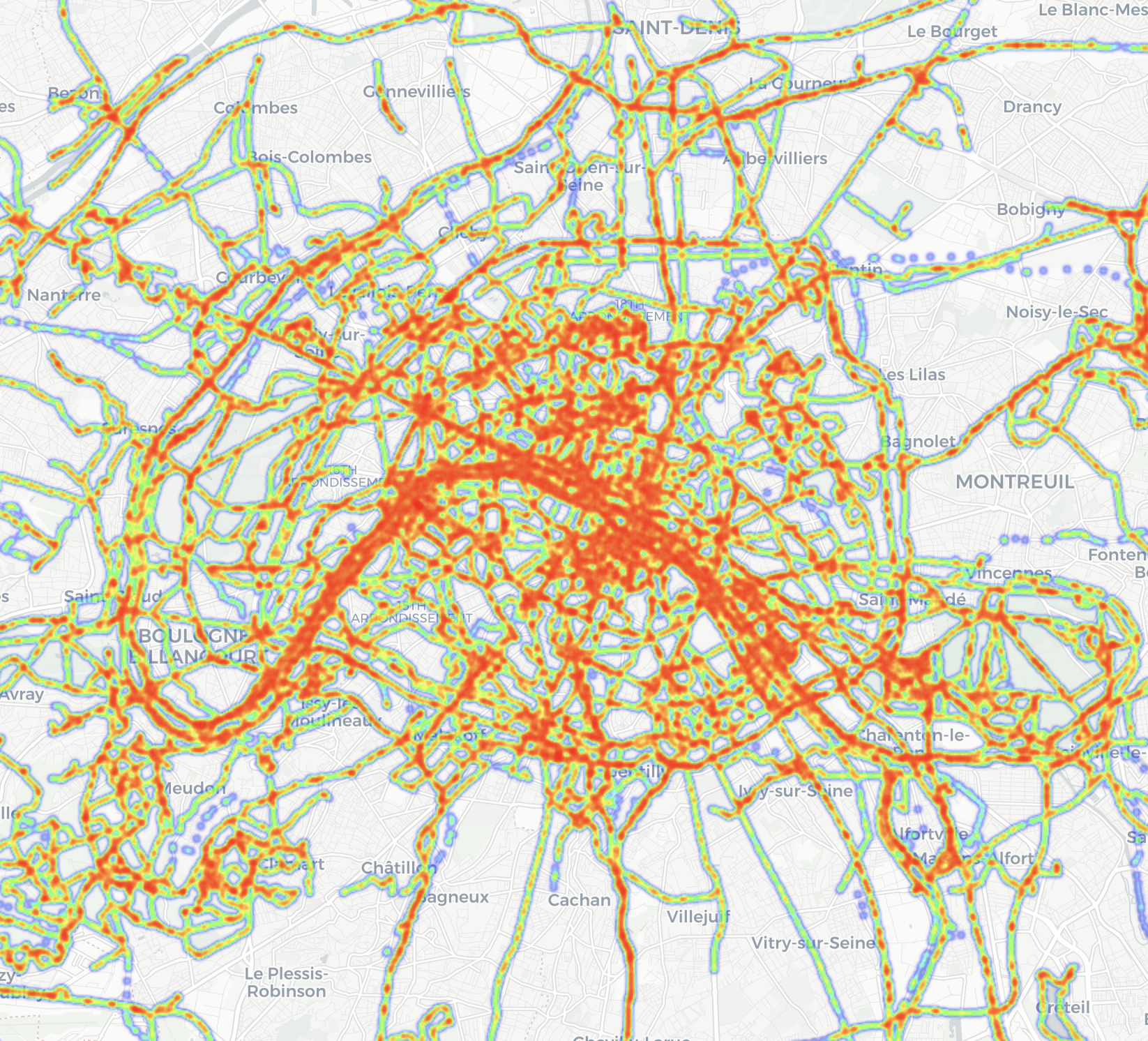}
    \caption{Heatmaps of NYC (\textit{left}) and Paris (\textit{right}) trajectories.}
    \label{fig:heatmaps}
\end{figure}

\vspace{0.2em}
\noindent \textbf{Raw GPS trajectory data.}
Raw GPS trajectory data are sourced from OpenStreetMap\footnote{Data available under the Open Database License (ODbL) (\url{https://opendatacommons.org/licenses/odbl/1.0/})}
by selecting geographic bounding boxes over Paris and New York City
and using the JOSM tool \cite{josm} along with the OSM API. Due to the very large number of GPS traces available in New York City, we first partition its bounding box into a grid, retrieve data from individual grid cells, and subsequently merge the data. At this stage, we also perform some basic preprocessing, mainly to reduce data size: we discard (1) duplicated trajectories, (2) trajectories for which no temporal information is available, and (3) trajectories that are not associated with a specific user-ID. In OSM, each user account can have multiple trajectories; accordingly, for each user, we chose to merge all their trajectories, hence yielding one trajectory per user.
In the end, for the city of Paris we obtain 1185 trajectories spanning August 2007 - February 2025, while for New York City we obtain 153180 trajectories spanning February 2008 - April 2025. Spatial distributions of these trajectories are visualized in Figure~\ref{fig:heatmaps}, focusing on high-density central areas for clarity. Timestamp conversions from UTC to the respective city time zones (Paris and New York City) are also performed. The bounding box values and the scripts used for downloading and preprocessing the GPS trajectories are publicly available in our GitHub repository \cite{matdataset_github_2025}.

\vspace{0.2em}
\noindent \textbf{POI data.}
We download POI data from OSM via the OSM API (code available in our repository \cite{matdataset_github_2025}). To capture a wide range of different POIs, we consider 
the following macro-categories: 
\textit{amenity}, which encompasses a wide range of facilities and services, such as banks, schools, toilets, and hospitals; \textit{shop}, which covers retail establishments, including supermarkets, bakeries, and clothing stores; \textit{tourism}, which includes tourist attractions and related services, like museums, hotels, and viewpoints; \textit{historic}, which denotes sites of historical significance, such as castles, ruins, and monuments; and \textit{leisure}, which includes green spaces, recreation, sport facilities, and more.
The bounding boxes used to download the POI data are the same as those used to download the trajectories.
For the city of Paris, we download a total of 109079 POIs, while for New York City we download a total of 66577 POIs. 

\vspace{0.2em}
\noindent \textbf{Weather data.}
We download historical weather data from Meteostat \cite{meteostat} via their API (code in our GitHub repository \cite{matdataset_github_2025}).
We select the weather stations of Charles De Gaulle Airport for Paris and the John F. Kennedy Airport for New York City, assuming they represent conditions for the entire urban areas.
We also perform some preprocessing to prepare the weather data for semantic enrichment, i.e., information concerning temperature and inferred sky conditions (e.g., sunny, rainy).
Finally, the weather data has a daily frequency and spans January 1st 1990 - June 3rd 2025.

\subsection{Raw trajectory preprocessing}
\label{sec: trajectory preprocessing}

\begin{figure}[t]
\centering
\includegraphics[width=0.8\linewidth]{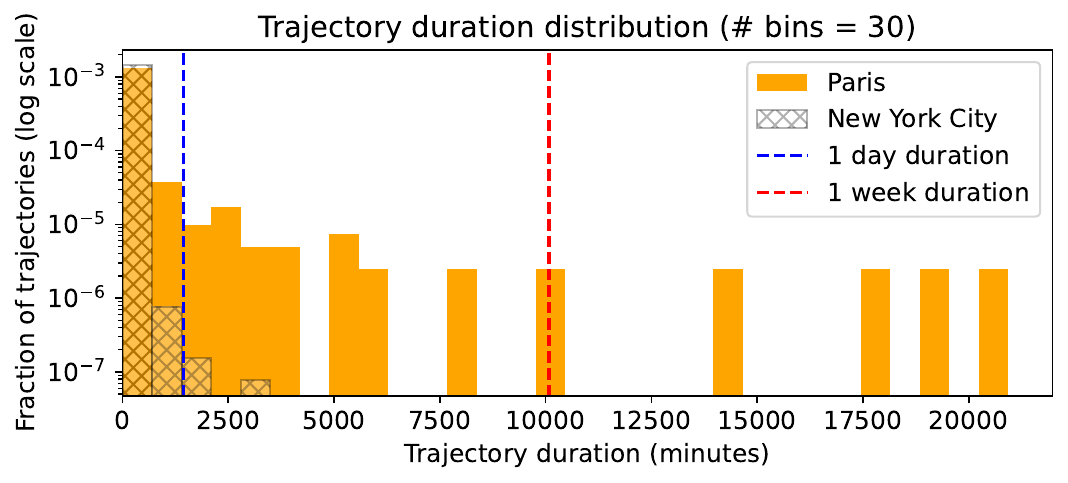}
\includegraphics[width=0.85\linewidth]{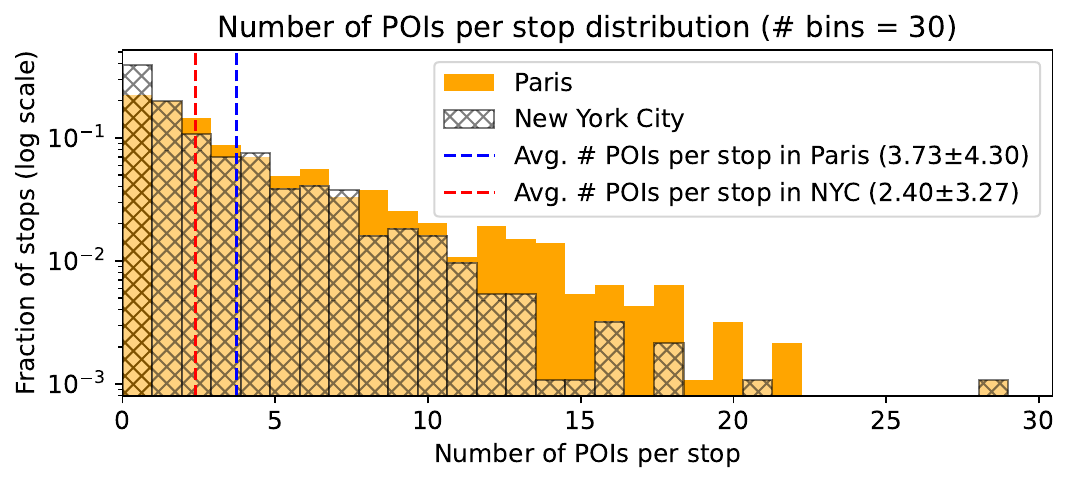}
\includegraphics[width=0.85\linewidth]{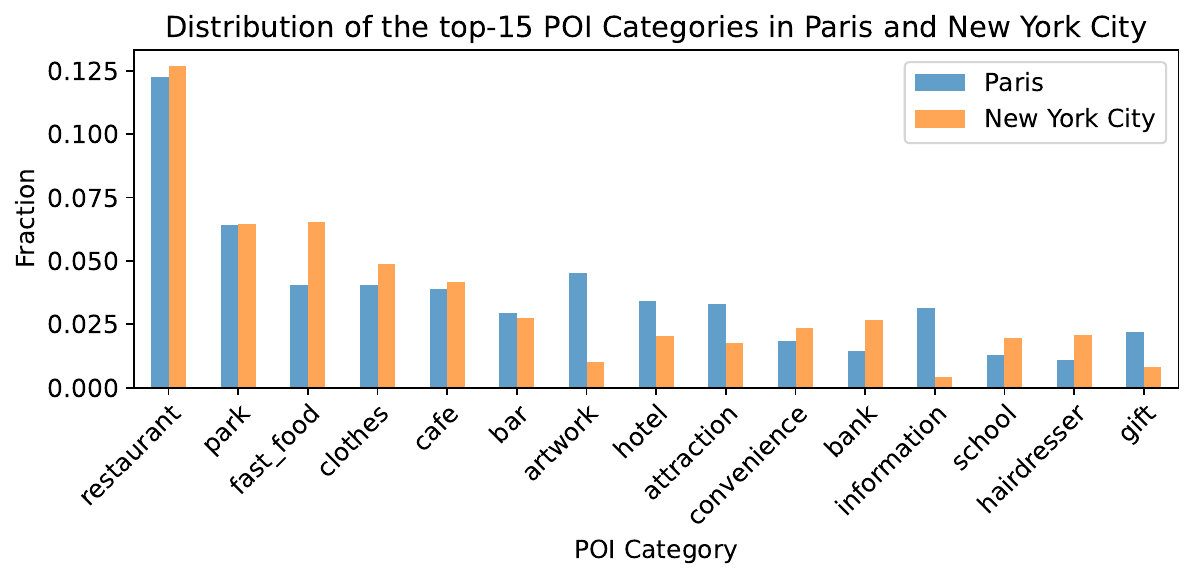}
\caption{Trajectory duration distribution (\textit{top plot}). Number of POIs per stop distribution (\textit{middle plot}). Distribution of the top-15 POI categories associated with stops (\textit{bottom plot}).}
\label{fig:dataset characteristics}
\end{figure}

Raw GPS trajectories from OpenStreetMap frequently exhibit issues such as very short duration, low sampling rates, or noisy data points. In the following, we report the main steps and criteria used to preprocess these trajectories.
First, we de-identify the user IDs, replacing each with a unique random number. We then require a minimum trajectory duration of 10 minutes, thus eliminating trajectories deemed too short to contain meaningful information. Finally, we require an average sampling rate of at least one sample every 2 minutes, hence removing trajectories considered too sparse.
\begin{table}[t]
  \centering
  \scriptsize
  \begin{tabular}{lcc}
  & \textbf{Paris} & \textbf{New York City}\\
  \toprule
    \textbf{Trajectories} & 581 & 18765\\
    \textbf{Trajectories with $\geq$ 4 weeks data} & 0 & 6\\
    \textbf{Trajectories with $\geq$ 1 week data} & 4 & 10\\
    \textbf{Trajectories with $\geq$ 1 day data} & 24 & 17\\
  \toprule
    \textbf{Sampling rate trajectories} & 13.5s $\pm$ 21.8s & 0.23s $\pm$ 2.10s\\
    \textbf{Duration trajectories} & 6h 22m 36s $\pm$ 1D 4h 15m 16s & 12m 48s $\pm$ 39m 41s\\
    \textbf{\# samples trajectories} & 2700 $\pm$ 4828 & 5617 $\pm$ 862\\
    \bottomrule
  \end{tabular}
   \caption{Metadata for the preprocessed raw trajectory data from Paris and New York City. \label{tab: osm dataset Paris}}
\end{table}
All these steps result in the generation of the two final preprocessed raw GPS trajectory datasets, whose characteristics are highlighted in Table \ref{tab: osm dataset Paris}\footnote{Time units are abbreviated as follows: D = day, h = hour, m = minute, s = second. Some values are reported as `mean $\pm$ standard deviation'}.
Each user in the datasets is associated with a single trajectory and an anonymous numerical identifier. The vast majority of trajectories last less than a day, and the temporal intervals spanned by Paris' trajectories are, on average, larger than New York City's, as also shown in Figure \ref{fig:dataset characteristics} (\textit{top plot}).

\subsection{Semantic Enrichment of Trajectories}
\label{sec: semantic enrichment}

Our semantic enrichment workflow comprises two main steps: (1) trajectory segmentation and (2) enrichment of trajectory segments with multiple semantic dimensions (as shown in Figure~\ref{fig:schema}). 
The tool we use to implement the workflow is MAT-Builder \cite{lettich2023semantic}. In the following, as well as in our public repositories \cite{matdataset_github_2025, zenodo_dataset2025}, we provide all the necessary details to understand, reproduce, and freely customize the semantic enrichment workflow.

\vspace{0.2em}
\noindent \textbf{Trajectory Segmentation.} 
This operation takes as input a dataset of preprocessed raw trajectories and partitions every trajectory into stop and move segments using the classic stop-and-move criterion \cite{spaccapietra2008conceptual}. A stop segment represents the event in which an individual is staying within a small area for at least some time. A move segment represents the transition from one stop to another. These segments represent the basic units for enriching the trajectories. Finally, we report that we set the minimum duration of a stop to 10 minutes,
while the maximum spatial radius a stop can have is set to 0.2 km. This yields 969 stops (mean duration: $193.96 \pm 833.72$ minutes) and 1479 moves (mean distance covered: $9.45 \pm 12.18$ km) for Paris, and 967 stops (mean duration: $36.31 \pm 110.87$ minutes) and 19474 moves (mean distance covered: $8.76 \pm 6.90$ km) for New York City.

\vspace{0.2em}
\noindent \textbf{Trajectory semantic enrichment.} This operation takes as input the segmented trajectories, along with the data necessary for enrichment with the various semantic dimensions. 
Stop segments are augmented by associating them with all points of interest (POIs) located within 50 meters of their centroids -- we consider this a reasonable distance to associate stops with potentially related POIs; we also report that this parameter is configurable in the workflow to suit specific requirements. Figure \ref{fig:dataset characteristics}, \textit{middle} and \textit{bottom} plots, reports some statistics concerning the POIs that have been associated with the stops. Overall, the stop segments have been enriched with a reasonable number of POIs on average (\textit{middle plot}), while the most frequent POI categories correspond to leisure-related activities (\textit{bottom plot}).
Next, the move segments are augmented with the \textit{inferred transportation mode}. Transportation modes are estimated using a random forest classifier trained on the GeoLife dataset, leveraging the kinematic features associated with each segment. The possible modes include ``walk", ``car", ``bike", ``bus", ``subway", and ``train".
Following this, trajectories are enriched with the \textit{weather} semantic dimension. Each trajectory is associated with the weather data (which we recall has a daily frequency and has been collected as detailed in Section \ref{sec: data collection}) that corresponds to the geographical area and time interval covered by the trajectory.
For example, a trajectory in Paris that spans $n$ distinct days is enriched with $n$ instances of the weather semantic dimension, each representing weather conditions for Paris on one of those $n$ days.

\vspace{-0.2cm}
\subsection{Synthetic social media data generation}
\label{sec: social media}
As shown in Figure~\ref{fig:schema}, the datasets are further enriched by adding a synthetic social media dimension. We opt to generate synthetic social media data, 
as linking trajectory users to their real social accounts is infeasible, and there would also be privacy concerns and API restrictions.
More specifically, we first select a state-of-the-art LLM, \textit{Llama-3.3-70B-Instruct}, which we use with half-precision (\textit{bfloat16}) to reduce computational costs. 
We then develop a custom prompting strategy to guide the model in producing plausible, realistic, and context-aware posts.

More precisely, posts are generated based on characteristics derived from the nearest POIs to trajectory stop segment centroids, primarily using POIs' names and categories as indicative context for plausible user activities.
Moreover, to introduce diversity and ensure balanced data, we use four synthetic user metadata categories: 
\textit{gender} (male, female, other), 
\textit{age} (18–24, 25–34, 35–44, 45–54, 55–64, 65+), 
\textit{ethnicity} (White, Black, Hispanic, Asian, Other), 
and \textit{social network} (Twitter, Instagram, Facebook, Tripadvisor).
The generation prompt explicitly instructs the model to behave as a \textit{creative social media post generator}, and that it must generate posts aligned to the information detailed above.
The prompt also instructs the model to avoid generic sentence starters, e.g., \textit{“I just visited”} or \textit{“Just spent”}, which were overrepresented in preliminary attempts. 
We also prompted the model to include hashtags.
For the generation parameters, we limit the output to 64 new tokens, set \texttt{do\_sample} to \texttt{true} to avoid deterministic outputs, and use a temperature value of $0.9$ to encourage diverse, yet coherent generations. 
Both the code and the prompt that generate social media data are publicly available on our GitHub repository  \cite{matdataset_github_2025}.

\subsection{Representing Enriched Trajectories}

To ensure usability and foster reproducible research, our datasets are released in two complementary formats: tabular and RDF.
The \textit{tabular representation} stores the enriched trajectories in several Pandas dataframes. For further details about this format, we refer the reader to our GitHub \cite{matdataset_github_2025} and Zenodo \cite{zenodo_dataset2025} repositories.

To support more advanced querying, semantic reasoning, and exploratory analysis of semantically enriched trajectories, we also provide an \textit{RDF-based knowledge graph representation}. In this format, entities such as users, trajectories, stops, moves, and POIs are explicitly modeled, along with the relationships among them. More details on the ontology we used to support  the Knowledge Graph representation can be found in \cite{lettich2023semantic} and in our repositories \cite{matdataset_github_2025, zenodo_dataset2025}.
Representing enriched trajectories in RDF enables SPARQL-based semantic queries and reasoning through triplestore technologies. Indeed, this transformation converts raw GPS traces into structured, context-aware mobility data that captures user routines, activities, movement patterns, and relationships with both the environment and social media.
Knowledge graphs offer a formal, machine-readable representation of spatial relationships, enabling both symbolic reasoning and embedding-based approaches. By explicitly linking spatial entities, they enhance the ability of large language models to perform targeted searches. Furthermore, this representation adheres to the FAIR principles \cite{fair_principles}.

\section{Potential Research Impacts}
\label{sec: task desc}

The resource \cite{matdataset_github_2025, zenodo_dataset2025} presented in this paper supports research in human behavior analysis, urban planning, and algorithm development. It is especially useful for model validation, knowledge graph construction, and foundation model design. Its open, customizable pipeline ensures long-term utility for the research community.
Spanning temporal, spatial, semantic, and social dimensions, its heterogeneous structure captures the complexity of urban mobility. The semi-synthetic nature makes it well-suited for experimental tasks such as movement prediction, behavioral modeling, and semantic reasoning.

\sloppypar{The raw trajectory data layer with weather information, makes it particularly well-suited for studies focused on mobility analysis in urban contexts, such as traffic prediction or urban movement modeling.
Building on this, the second layer of the dataset, which includes inferred semantic annotations as shown in Figure~\ref{fig:schema}, enables the development and evaluation of advanced methods in activity recognition, predictive modeling, and recommender systems, like personalized route for tourists. 
The inclusion of synthetic social media content opens opportunities for multi-modal analysis, especially methods that combine sentiment analysis with location or movement data.
To validate this assumption, we tested the generated posts with three models for sentiment analysis, i.e., \texttt{cardiffnlp/twitter-roberta-base-sentiment}, \texttt{nlptown/bert-base-multilingual-uncased-sentiment}, and \texttt{finiteautomata/bertweet-base-sentiment-analysis}, obtaining an accuracy score of 98.9\%, 98.5\%, and 97.0\%  among the whole generated posts. After this evaluation, we manually checked the posts not labelled as expected and observed that they were wrongly tagged, mostly due to some foreign words contained in the generated posts.
This makes the dataset highly suitable for testing algorithms that integrate social signals into mobility-aware applications.}

While the classical tabular format is well-suited for machine learning pipelines, 
the RDF representation adds significant value by enabling knowledge graph construction and semantic reasoning \cite{UrbanKG,MobilityPredictionKG}. It offers a rich framework for linking structured mobility data with contextual semantic layers, supporting advanced and interpretable analyses like semantic enrichment, spatiotemporal reasoning, user-POI linking, and natural language processing for geographical data. 
This aspect could be useful in combination with LLMs to capture spatial semantics implicitly from large text corpora. Combining these approaches holds strong potential: knowledge graphs provide a reliable spatial structure, while LLMs bring powerful natural language understanding and generation capabilities.
For instance, it can fuel emerging research on chatbot architectures capable of answering complex, natural language questions about mobility data \cite{padoan2024mobility, chen2024travelagent, devunuri2024chatgpt}. These systems often rely on multiple specialized agents -- each driven by an LLM -- that collaboratively query diverse data sources and process the retrieved information to formulate answers. Recent studies have already demonstrated the automatic generation of SPARQL queries over knowledge graphs \cite{kovriguina2023sparqlgen}, which could be directly applied to the semantic layers of our dataset. Another promising role for this resource is in the context of foundation models and LLMs. 
By publicly releasing these datasets  and the building pipeline, we aim to empower the CIKM community with a resource that enables experiments and testing of methods and algorithms 
for more interpretable, explainable, and context-aware approaches to human behavior analysis.

\section*{Acknowledgments}

This research has been partially funded by the European Union’s Horizon Europe research and innovation program EFRA (Grant Agreement Number 101093026). 
This work was also supported by the MUSIT Project through the European Union’s Horizon 2020 research and innovation program under Marie-Sklodowska Curie grant agreement no. 101182585. 
Views and opinions expressed are however those of the authors only and do not necessarily reflect those of the European Union or European Commission-EU. Neither the European Union nor the granting authority can be held responsible for them.

\bibliographystyle{abbrvurl}
\bibliography{biblio}

@article{MobilityPredictionKG,
author = {Yu, Qiaohong and Wang, Huandong and Liu, Yu and Jin, Depeng and Li, Yong and Zhu, Lin and Feng, Junlan},
title = {Mobility Prediction via Rule-enhanced Knowledge Graph},
year = {2024},
issue_date = {November 2024},
publisher = {Association for Computing Machinery},
address = {New York, NY, USA},
volume = {18},
number = {9},
issn = {1556-4681},
url = {https://doi.org/10.1145/3677019},
doi = {10.1145/3677019},
journal = {ACM Trans. Knowl. Discov. Data},
month = oct,
articleno = {215},
numpages = {21}
}

@article{UrbanKG,
author = {Liu, Yu and Ding, Jingtao and Fu, Yanjie and Li, Yong},
title = {UrbanKG: An Urban Knowledge Graph System},
year = {2023},
issue_date = {August 2023},
publisher = {Association for Computing Machinery},
address = {New York, NY, USA},
volume = {14},
number = {4},
issn = {2157-6904},
url = {https://doi.org/10.1145/3588577},
doi = {10.1145/3588577},
journal = {ACM Trans. Intell. Syst. Technol.},
articleno = {60},
numpages = {25}
}

@article{KappSurvey24,
author = {Kapp, Alexandra and Hansmeyer, Julia and Mihaljevi\'{c}, Helena},
title = {Generative Models for Synthetic Urban Mobility Data: A Systematic Literature Review},
year = {2023},
issue_date = {April 2024},
publisher = {Association for Computing Machinery},
address = {New York, NY, USA},
volume = {56},
number = {4},
issn = {0360-0300},
url = {https://doi.org/10.1145/3610224},
doi = {10.1145/3610224},
journal = {ACM Comput. Surv.},
month = nov,
articleno = {93},
numpages = {37}
}

@manual{zheng2011geolife,
author = {Zheng, Yu and Fu, Hao and Xie, Xing and Ma, Wei-Ying and Li, Quannan},
title = {Geolife GPS trajectory dataset - User Guide},
year = {2011},
month = {July},
url = {https://www.microsoft.com/en-us/research/publication/geolife-gps-trajectory-dataset-user-guide/},
edition = {Geolife GPS trajectories 1.1},
}

@inproceedings{GoceData24,
  author       = {Joseph Zuber and
                  Xu Teng and
                  Andreas Z{\"{u}}fle and
                  Goce Trajcevski},
  title        = {Data and Resources for Combining Point of Interest Semantics, Locations,
                  and Road Networks},
  booktitle    = {Proceedings of the 32nd {ACM} International Conference on Advances
                  in Geographic Information Systems, {SIGSPATIAL} 2024, Atlanta, GA,
                  USA, 29 October 2024 - 1 November 2024},
  pages        = {705--708},
  publisher    = {{ACM}},
  year         = {2024},
  url          = {https://doi.org/10.1145/3678717.3691300},
  doi          = {10.1145/3678717.3691300},
  bibsource    = {dblp computer science bibliography, https://dblp.org}
}

@inproceedings{AndreasPatterns24,
  author       = {Hossein Amiri and
                  Will Kohn and
                  Shiyang Ruan and
                  Joon{-}Seok Kim and
                  Hamdi Kavak and
                  Andrew T. Crooks and
                  Dieter Pfoser and
                  Carola Wenk and
                  Andreas Z{\"{u}}fle},
  title        = {The Patterns of Life Human Mobility Simulation},
  booktitle    = {Proceedings of the 32nd {ACM} International Conference on Advances
                  in Geographic Information Systems, {SIGSPATIAL} 2024, Atlanta, GA,
                  USA, 29 October 2024 - 1 November 2024},
  pages        = {653--656},
  publisher    = {{ACM}},
  year         = {2024},
  url          = {https://doi.org/10.1145/3678717.3691319},
  doi          = {10.1145/3678717.3691319},
  bibsource    = {dblp computer science bibliography, https://dblp.org}
}

@inproceedings{kovriguina2023sparqlgen,
  title={SPARQLGEN: One-Shot Prompt-based Approach for SPARQL Query Generation.},
  author={Kovriguina, Liubov and Teucher, Roman and Radyush, Daniil and Mouromtsev, Dmitry},
  booktitle={SEMANTiCS (Posters \& Demos)},
  year={2023}
}

@article{lettich2023semantic,
  title={Semantic enrichment of mobility data: A comprehensive methodology and the mat-builder system},
  author={Lettich, Francesco and Pugliese, Chiara and Renso, Chiara and Pinelli, Fabio},
  journal={IEEE Access},
  volume={11},
  pages={90857--90875},
  year={2023},
  publisher={IEEE}
}

@article{spaccapietra2008conceptual,
title = {A conceptual view on trajectories},
journal = {Data \& Knowledge Engineering},
volume = {65},
number = {1},
pages = {126-146},
year = {2008},
issn = {0169-023X},
doi = {https://doi.org/10.1016/j.datak.2007.10.008},
author = {Stefano Spaccapietra and Christine Parent and Maria Luisa Damiani and Jose Antonio {de Macedo} and Fabio Porto and Christelle Vangenot}
}

@inproceedings{padoan2024mobility,
  title={Mobility ChatBot: supporting decision making in mobility data with chatbots},
  author={Padoan, Lorenzo and Cesetti, Margherita and Brunello, Luca and Antonelli, Marco and Zamengo, Bruno and Silvestri, Francesco},
  booktitle={2024 25th IEEE International Conference on Mobile Data Management (MDM)},
  pages={295--300},
  year={2024},
  organization={IEEE}
}

@article{Monreale23,
  author       = {Anna Monreale and
                  Roberto Pellungrini},
  title        = {A Survey on Privacy in Human Mobility},
  journal      = {Trans. Data Priv.},
  volume       = {16},
  number       = {1},
  pages        = {51--82},
  year         = {2023},
  url          = {http://www.tdp.cat/issues21/abs.a464a22.php},
  bibsource    = {dblp computer science bibliography, https://dblp.org}
}

@article{Gomes24,
  author       = {Fernanda Oliveira Gomes and
                  Roberto Pellungrini and
                  Anna Monreale and
                  Chiara Renso and
                  Jean Everson Martina},
  title        = {TrajectGuard: {A} Comprehensive Privacy-Risk Framework for Multiple-Aspects
                  Trajectories},
  journal      = {{IEEE} Access},
  volume       = {12},
  pages        = {136354--136378},
  year         = {2024},
  url          = {https://doi.org/10.1109/ACCESS.2024.3462088},
  doi          = {10.1109/ACCESS.2024.3462088},
  bibsource    = {dblp computer science bibliography, https://dblp.org}
}

@article{devunuri2024chatgpt,
  title={Chat{GPT} for {GTFS}: benchmarking {LLM}s on {GTFS} semantics... and retrieval},
  author={Devunuri, Saipraneeth and Qiam, Shirin and Lehe, Lewis J},
  journal={Public Transport},
  volume={16},
  number={2},
  pages={333--357},
  year={2024},
  publisher={Springer}
}

@article{chen2024travelagent,
  title={Travel{A}gent: An {AI} assistant for personalized travel planning},
  author={Chen, Aili and Ge, Xuyang and Fu, Ziquan and Xiao, Yanghua and Chen, Jiangjie},
  journal={arXiv preprint arXiv:2409.08069},
  year={2024}
}

@book{matsim2016,
  title={The multi-agent transport simulation MATSim},
  author={W Axhausen, Kay and Horni, Andreas and Nagel, Kai},
  year={2016},
  publisher={Ubiquity Press}
}

@misc{zenodo_dataset2025,
  author       = {Pugliese, Chiara and
                  Lettich, Francesco and
                  Rocchietti, Guido and
                  Renso, Chiara and
                  Pinelli, Fabio},
  title        = {Dataset repository for the  resource paper "{A
                   Semantically Enriched Mobility Dataset with
                   Contextual and Social Dimensions}"
                  },
  month        = jun,
  year         = 2025,
  publisher    = {Zenodo},
  doi          = {10.5281/zenodo.15624419},
  url          = {https://doi.org/10.5281/zenodo.15624419},
}

@misc{matdataset_github_2025,
  title        = {GitHub repository containing the code for the "{Human Mobility Datasets Enriched
With Contextual and Social Dimensions}" resource paper},
  url          = {https://github.com/Fr4nz83/MAT-Dataset}
}

@misc{taxiporto2015,
  title        = {{ECML/PKDD 15}: Taxi Trajectory Prediction Challenge},
  year         = {2015},
  url          = {https://www.kaggle.com/c/pkdd-15-predict-taxi-service-trajectory-i/overview},
  note = {Accessed on: 2025-06-11}
}

@misc{josm,
  title        = {{Java OpenStreetMap (JOSM)} editor website},
  url          = {https://josm.openstreetmap.de/},
  note = {Accessed on: 2025-06-11}
}

@misc{osm,
  title        = {{OpenStreetMap} website},
  url          = {https://www.openstreetmap.org/},
  note = {Accessed on: 2025-06-11}
}

@misc{meteostat,
  title        = {Meteostat website},
  url          = {https://meteostat.net/},
  note = {Accessed on: 2025-06-11}
}

@inproceedings{yang2019revisiting,
  title={Revisiting user mobility and social relationships in lbsns: a hypergraph embedding approach},
  author={Yang, Dingqi and Qu, Bingqing and Yang, Jie and Cudre-Mauroux, Philippe},
  booktitle={The world wide web conference},
  pages={2147--2157},
  year={2019}
}

@article{fair_principles,
    author = {Jacobsen, Annika and de Miranda Azevedo, Ricardo and Juty, Nick and et al.},
    title = {FAIR Principles: Interpretations and Implementation
 Considerations},
    journal = {Data Intelligence},
    volume = {2},
    number = {1-2},
    pages = {10-29},
    year = {2020},
    month = {01},
    issn = {2641-435X},
    doi = {10.1162/dint_r_00024}
}

\end{document}